\documentclass[letterpaper, 10 pt, conference]{ieeeconf}  

\IEEEoverridecommandlockouts                              
\usepackage{optidef}
\usepackage{graphicx} 
\usepackage{times} 
\usepackage{amsmath} 
\usepackage{amssymb}  
\usepackage[hidelinks]{hyperref}
\usepackage[printonlyused]{acronym}
\usepackage{nomencl}
\usepackage{mathrsfs}
\usepackage[caption=false,font=footnotesize]{subfig}
\newcommand{\norm}[1]{\left\lVert#1\right\rVert}
\usepackage{makecell}
\usepackage{longtable}
\usepackage{cite}

\begin{document}
\bstctlcite{IEEEexample:BSTcontrol}
\title{\LARGE\bf{Towards Scalable Continuous-Time Trajectory Optimization for Multi-Robot Navigation}}
%
%
%

\author{Shravan Krishnan, Govind Aadithya Rajagopalan, Sivanathan Kandhasamy and Madhavan Shanmugavel
\thanks{The authors are with Autonomous Systems Lab, SRM Institute of Science and Technology, India  
}
\thanks{This work was funded by SRM Institute of Science and Technology, India  
}
\thanks{Corresponding Author: Shravan Krishnan, 
shravan\_krishnan@srmuniv.edu.in}
}

\maketitle

\begin{abstract}
Scalable multi-robot transition is essential for ubiquitous adoption of robots. As a step towards it, a computationally efficient decentralized algorithm for continuous-time trajectory optimization in multi-robot scenarios based upon model predictive control is introduced. The robots communicate only their current states and goals rather than sharing their whole trajectory; using this data each robot predicts a continuous-time trajectory for every other robot exploiting optimal control based motion primitives that are corrected for spatial inter-robot interactions using least squares. A non linear program (NLP) is formulated for collision avoidance with the predicted trajectories of other robots. The NLP is condensed by using time as a parametrization resulting in an unconstrained optimization problem and can be solved in a fast and efficient manner. Additionally, the algorithm resizes the robot to  accommodate it's trajectory tracking error. The algorithm was tested in simulations on Gazebo with aerial robots. Early results indicate that the proposed algorithm is efficient for upto forty homogeneous robots and twenty one heterogeneous robots occupying 20\% of the available space.
\end{abstract}

\section{Introduction}

Generation of collision-free trajectories for multiple robots in a shared workspace is a critical necessity for robots to become imbibed into the daily ecosystem. To safely allow such an entrenchment, multiple menages of methods have been developed.

One such family of methods is sampling based motion planning \cite{solovey2016finding,bekris2012safe,yu2018effective,yu2016optimal}. Sampling based motion planning algorithms generate geometric paths by incrementally exploring the available space. Such methods are restricted to first order systems but can guarantee safety and global optimality. Unfortunately, these methods suffer from the curse of dimensionality and are difficult to scale to large number of robots and higher order dynamics.

To mitigate the scalability problems with regards to higher order dynamics and high dimensional spaces, an increasing number of methods have been proposed wherein collision-free geometric paths are initially planned and then refined using polynomial trajectory optimization techniques  \cite{tang2018hold,honig2018,debord2018trajectory}.

At the other end of the spectrum, direct optimal control/optimization based methods have been proposed that directly attempt to plan collision-free trajectories \cite{agualiro2012scp,luis2019trajectory,chen2015scp,hamer2018fast,robinson2018efficient}. These methods are capable of handling higher dimensional spaces directly but at the cost of  greater number of constraints on the NLP, discrete collision-checking and local trajectory optimization. Consequently, such methods can be parallelized and extended to use on-demand collision avoidance and pair-wise collision detection to generate trajectories for 200 three dimensional aerial robots in a few seconds \cite{luis2019trajectory,hamer2018fast}.

The above presented methods- with the exception of \cite{bekris2012safe}- are offline methods and computationally infeasible for a real-time implementation. Offline methods are reasonable in entertainment, but real world implementations require resilience.  Thus, integrating robots into the daily ecosystem is not practical using these algorithms wherein the environment itself is constantly evolving and therefore requires replanning trajectories. Also, for an efficient real-time replanning, decentralizing the trajectory generation problem is more efficient and averse to failure.

One such family of decentralized collision avoidance algorithms is velocity obstacles \cite{fiorini1998}. In such techniques, trajectories are planned in the velocity space \cite{Berg2008}. These methods have also been extended to higher order dynamics \cite{rufli2013reciprocal} and for high speed avoidance \cite{liu2018avoidance}. Detrimentally, these methods require a reference velocity to be known.

Another paradigm for collision avoidance is the usage of safe regions. These safe regions can be formulated by using buffered voronoi cells \cite{2017fast,csenbacslar2019robust} or built using separating hyperplanes \cite{tang2018hold,honig2018}. Buffered voronoi cells have been implemented in a decentralized fashion by relying only on the moving obstacles' current position. This is sub-optimal when dealing with robots moving at higher velocities.

Safety barrier certificates is another branch of cooperative collision avoidance wherein a stabilizing controller is reshaped for collision avoidance using barrier certificates \cite{safe2017certificate,wang2017safety}. 

\begin{figure*}[h!]
\centering
\subfloat[][]{\includegraphics[width=0.3\textwidth]{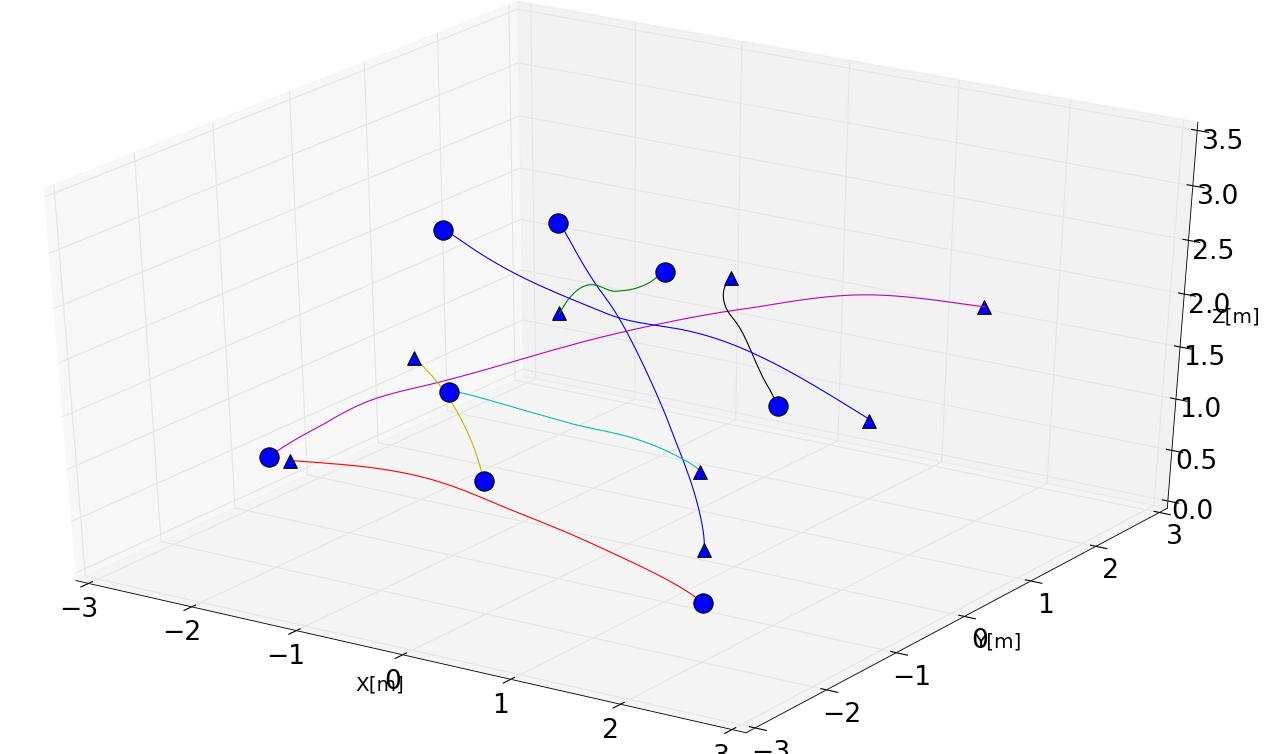}\label{8_robot}} 
\hspace{2mm}
\subfloat[][]{\includegraphics[width=0.32\textwidth]{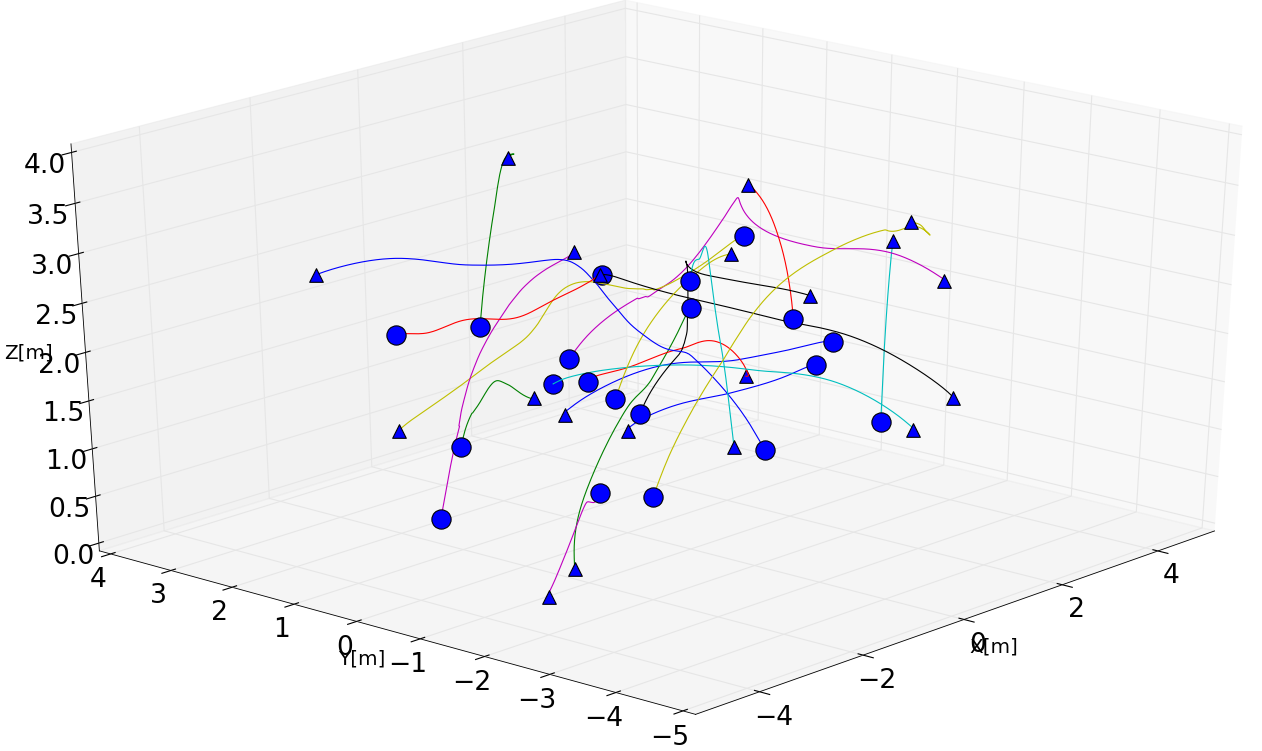}\label{20_robot}}
\hspace{2mm}
\subfloat[][]{\includegraphics[width=0.32\textwidth]{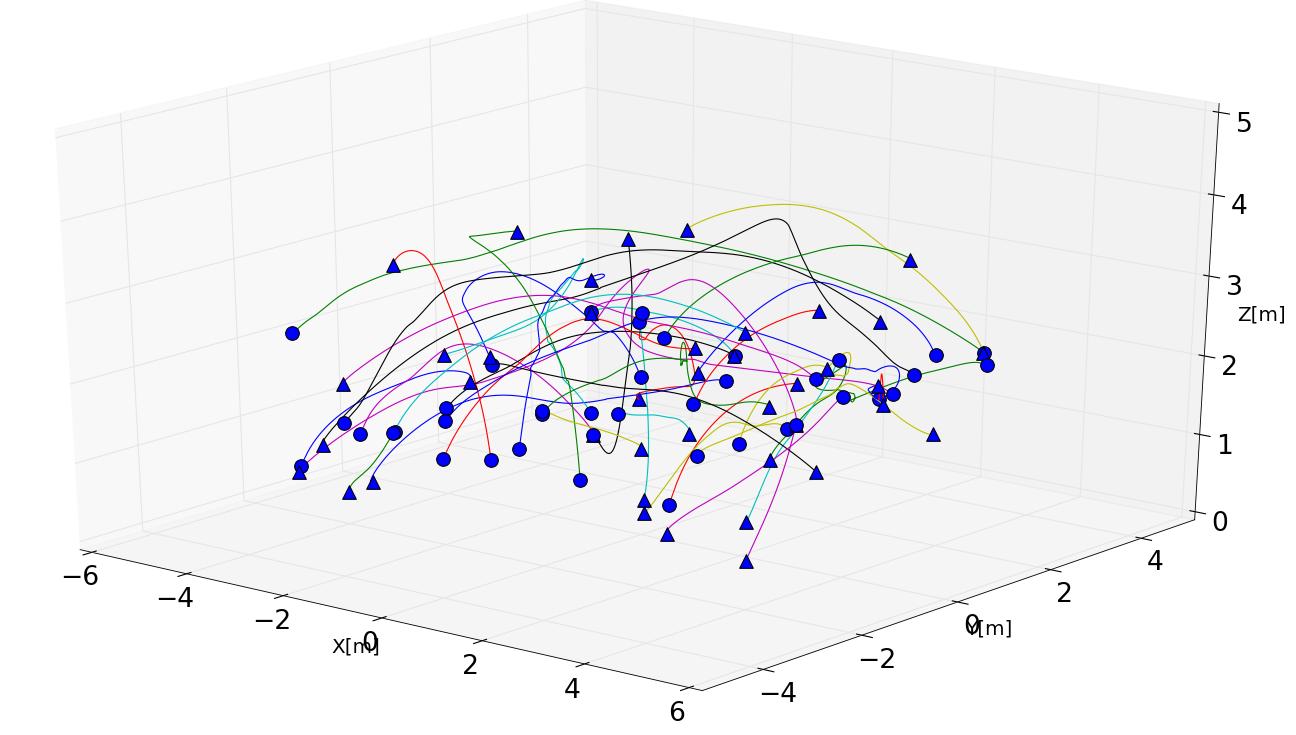}\label{40_robot}}
\caption{The executed trajectories as robots traverse across an environment. The coloured circles indicate the starting position of the robots while the triangles indicate the end position. Fig \ref{8_robot} shows the trajectory traversed for eight robots,Fig \ref{20_robot} for twenty robots and Fig \ref{40_robot} the trajectories for forty robots. Video of simulations is available at \href{https://bit.ly/2krAI2h}{https://bit.ly/2krAI2h}}
\label{Generatedtrajectories}
\end{figure*}

Additionally, as the problem becomes decentralized, accounting for the trajectories of other robots/moving-obstacles becomes a problem. This is often neglected in literature by assuming the robots' share their trajectories or having constant acceleration or velocity assumptions. Moreover, not sharing trajectories also alleviates privacy concerns over the system.

Stemming from these deficiencies, with a focus on improving real-time scalability, a decentralized local trajectory optimization algorithm that integrates a layer to predict trajectories for other robots is presented. The prediction is done by using a computationally efficient method to generate a minimum-time trajectories for other robots. This trajectory is further reshaped spatially to account for interactions by using a moving horizon of the robots' past states.

Using the reshaped trajectories, a collision-free trajectory is planned consisting of a NLP that minimizes time and integral of jerk squared. Additionally, exponential barrier functions are used to augment collision avoidance and dynamic limits.  The usage of soft constraints is guided by it allowing for a continuous time verification and if necessary, can be violated by small margins. The NLP is condensed using time as a free variable. The resulting unconstrained optimization problem's solution denotes the duration of the trajectory and based upon it, polynomial coefficients of the trajectory are constructed. 

Keeping with the receding horizon principle, a part of the trajectory is executed and then overall trajectory is replanned.

A recent work \cite{Krishnan2019} attempted a similar problem as the one presented in this work but, (i) Only accounted for spherical Robots, (ii) Solved a constrained optimization problem directly, (iii) The prediction did not account for interactions.

Moreover, in this work, the robots are assumed to imperfectly track the trajectories. However, rather than triggering replanning upon such an event \cite{Luis2019online}, trajectory tracking errors are accounted for in the obstacle avoidance.

The contributions of this work are:
\begin{itemize}
    \item A computationally efficient trajectory prediction algorithm that spatially accounts for inter-robot interactions
    \item An online trajectory replanning algorithm for multiple robots accounting for imperfect trajectory tracking
\end{itemize}

The rest of the paper is organized as: Section \ref{system model} explains system model, Section \ref{prediction} showcases the Trajectory prediction method. Section \ref{traj optimization} details the trajectory optimization problem. Section \ref{results} details the simulations of the algorithm  and Section \ref{conclusion} concludes the paper.

\section{System Model}
\label{system model}
Each of the robots is modeled as a third order integrator moving in $N$ dimensions. The system is represented with state $\textbf{x}=[p_1 \hspace{1mm} v_1 \hspace{1mm} a_1 \hdots a_N]$ with $p$ being position, $v$ being velocity and $a$ being acceleration with  input $u=[j_1 \hdots j_n]$ where $j$ is jerk. The dynamics- in each dimension- is:
\begin{equation}
\dot{\textbf{x}}(t)=[v(t) \hspace{1mm} a(t) \hspace{1mm} j(t)]
\label{dynam}
\end{equation}

The selection of a third order integrator is motivated by two factors. One, the differential flatness of a variety of mobile robots \cite{mellinger2011minimum,optimalwalambe2016}; two, measuring higher derivative beyond acceleration is intractable. Additionally, the robots share their current state $\textbf{x}_0$, desired pose $p_{end}$ and size. 

\subsection{Accommodating Collision Avoidance}
 Collision avoidance is accounted for in a pair-wise basis. Additionally, in the case of aerial robots, downwash leads to collisions \cite{tang2018hold}. Therefore, building on \cite{luis2019trajectory,honig2018} the robots are modeled as $N$ dimensional axis-aligned prolate spheroids. Furthermore, an additional term is incorporated in each axis to account for the imperfections in trajectory tracking. Thus, the overall collision avoidance constraint is:
\begin{equation}
    d(p(t),p_{obs}(t))=\sum_{i=1}^N \frac{(p_i(t) - p_{obs}(t))^2}{(r_i + \xi)^2}> 1
    \label{collisionconstraint}
\end{equation}

With $\xi$ being the trajectory tracking error and $r_i$ being the appropriate radii in each dimension. In the case of $N$ =3, $r_1 = r_2 $ and $r_3= \eta r_1 $ with $\eta > 1$. So the size of the robots is denoted by $r_1$.  
\subsection{Accounting Trajectory Tracking Error}
There has been a few attempts to calculate the error in tracking trajectories. Fixing a maximal error offline before online trajectory optimization is tried in \cite{mora2018cooperative}. This requires the errors to be known \textit{a priori}. In  \cite{kousik2019safe}, a forward reachable set was calculated before executing the trajectory which is computationally expensive to perform online.  


\begin{figure*}
\centering
\subfloat[][]{\includegraphics[width=0.29\textwidth]{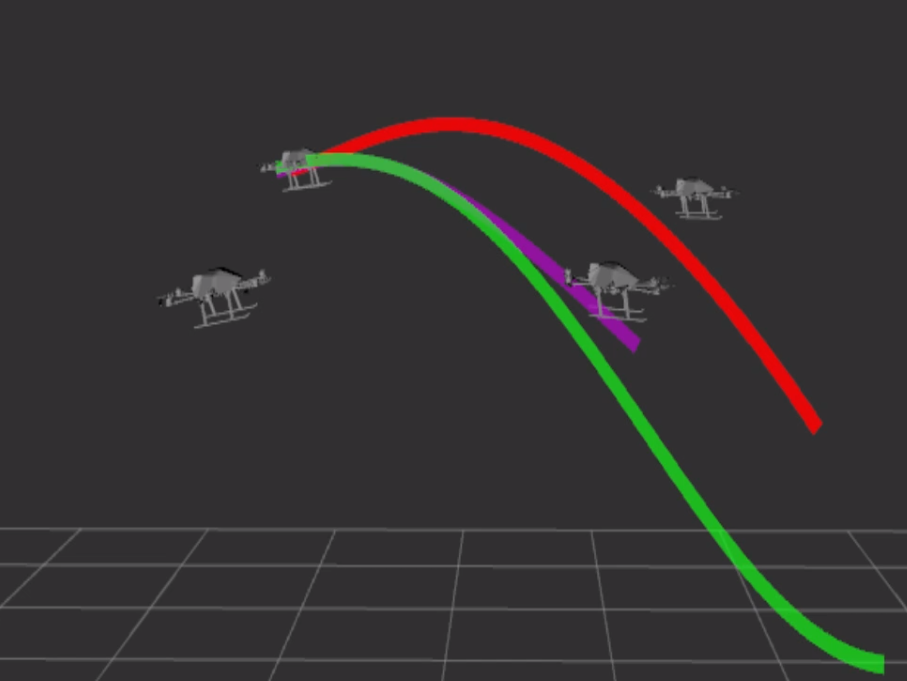}\label{4_predict}} 
\hspace{2mm} \subfloat[][]{\includegraphics[width=0.31\textwidth]{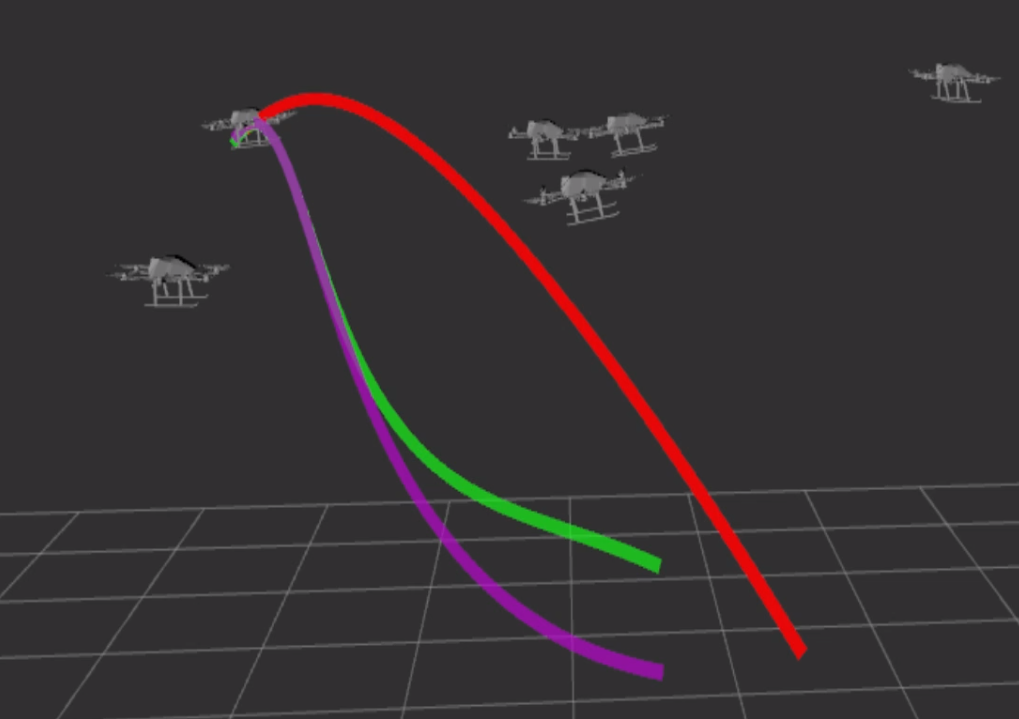}\label{6_predict}}\hspace{2mm}
\subfloat[][]{\includegraphics[width=0.34\textwidth]{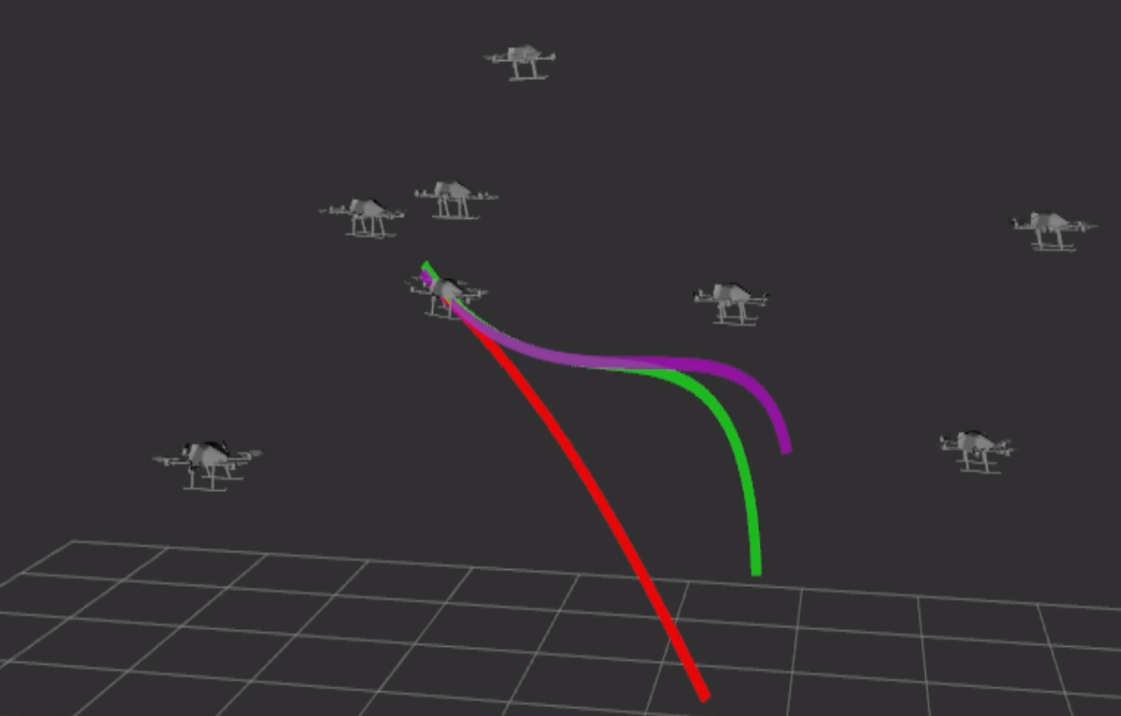}\label{8_predict}}
\caption{The deviation between the predicted trajectories and optimized collision-free trajectory with increasing number of robots. Red trajectory is the optimized trajectory, lilac the trajectory from the method in Section \ref{jerk-smooth} and green the prediction method detailed in Section \ref{prediction}.  Fig \ref{4_predict} shows prediction with four robots. Fig \ref{6_predict} shows prediction with six robots. Fig \ref{8_predict} shows prediction with eight robots. The predicted trajectories are from the point of view of an other robot}
\label{visualise prediction}
\end{figure*}

This work attempts to calculate the trajectory tracking error online using a moving horizon of weighted root mean square between planned trajectory and actual state discretely. That is:
\begin{equation}
    \xi = \sqrt{\frac{\sum_{i=1}^K (\textbf{x}_i-\Tilde{\textbf{x}}_i)^T Q (\textbf{x}_i-\Tilde{\textbf{x}}_i)}{K} }
\end{equation}

With, $K$ being the horizon steps in the moving horizon, $\Tilde{\textbf{x}}_i$ being $i$\textsuperscript{th} observed state. $Q$ is a positive semi definite matrix that assigns weights to tracking errors of position,velocity and acceleration. This is computationally efficient and easily computable online but ignores trajectory dependent tracking errors. 

\section{Trajectory Prediction}
\label{prediction}

The trajectories are predicted in a sequence consisting of two steps. In the first
step, jerk-smooth trajectories are generated for the linear quadratic minimum time problem \cite{verriest1991linear} using an extension of the closed-form solutions for motion primitive generation for multirotors \cite{efficient2015quad}. The usage of trajectory duration as a free variable allows an extra degree of freedom in the generated trajectory rather than fixing a duration arbitrarily. The second stage of the algorithm takes interactions between the robots into account via reshaping of the previous step's trajectory by solving a constrained least squares problem with regards to the jerk-smooth trajectory and the actual robot trajectory.


\subsection{Generating Jerk-smooth Trajectories}
\label{jerk-smooth}
A smooth trajectory is generated by minimizing:
\begin{argmini}
{}{\int_{0}^{T}  \norm{u} ^2 +1 \hspace{1mm} dt}{}{}
\label{pmpobjective}
\end{argmini}

Here, $\norm{.}$ is the euclidean norm, $u$ is jerk. The trajectory duration $T$ couples the objective in each dimension. For ease of exposition, the trajectory is shown for a single dimension. The trajectory is parameterized by the current state and  the (partially defined) end state \footnote{The system  is linear and has a convex objective}. 

The Hamiltonian of the third order system with the objective from Eq.\eqref{pmpobjective} is:

\begin{equation}
H(x,\lambda,u)= \norm{u}^2 +1 + \lambda_1v + \lambda_1a + \lambda_3u
\label{hamiltonian}
\end{equation}




Solving Eq.\eqref{hamiltonian} using Pontygarin's maximum principle \cite{dynamicprogramming}  yields the state trajectory:

\begin{equation}
\begin{split}
p^*(t)&= \frac{\beta_1t^5}{120}+\frac{\beta_2t^4}{24}+\frac{\beta_3t^3}{6}+\frac{a_0t^2}{2}+v_0t+p_0 \\
v^*(t)&=\frac{\beta_1t^4}{24}+\frac{\beta_2t^3}{6}+\frac{\beta_3t^2}{2}+a_0t+v_0 \\
a^*(t)&=\frac{\beta_1t^3}{6}+\frac{\beta_2t^2}{2}+\beta_3t+a_0
\end{split}
\label{optimal_state}
\end{equation}

with costate equation:

\begin{equation}
\Lambda^*(t)=-\begin{bmatrix}
2\beta_1 \\
2\beta_1t+ 2\beta_2 \\
\beta_1t^2+ 2\beta_2t + 2\beta_3
\end{bmatrix}
\label{costatesol}
\end{equation}

Position is the only end state that has to be constrained. This is substituted into Eq. \eqref{optimal_state} resulting in 
\begin{equation}
p_{end}(T)= \frac{\beta_1T^5}{120}+\frac{\beta_2T^4}{24}+\frac{\beta_3T^3}{6}+\frac{a_0T^2}{2}+v_0T+p_0 
\end{equation}

The three unknown coefficients($\beta_1,\beta_2,\beta_3$) are solved such that the corresponding costates will be zero at the free states: velocity and acceleration. \cite{dynamicprogramming}. This leads to a linear system from Eq. \eqref{costatesol} with the three unknown coefficients. Coincidentally, this method allows representing the unknown coefficients as functions parameterized by trajectory duration $T$, known initial states $\textbf{x}_0$ and end position $p_{end}$. Resulting in:

\begin{equation}
\begin{bmatrix}
\beta_1 \\
\beta_2 \\
\beta_3 \\
\end{bmatrix} = \frac{1}{T^5}\begin{bmatrix}
20 \\
-20T \\
10T^2
\end{bmatrix}(p_{end}-(p_0+p_0T+ \frac{a_0T^2}{a}))
\label{equation}
\end{equation}
Furthermore, due to the trajectory duration $T$ being a variable to optimize, $H(x,\lambda,u)$ of the system is now zero instead of a constant \cite{dynamicprogramming}. Hence,

\begin{equation}
H(x,\lambda,u)= \norm{u}^2 +1 + \lambda_1v + \lambda_1a + \lambda_3u=0
\label{Hamil}
\end{equation}

Substituting Eq. \eqref{equation} into Eq. \eqref{Hamil} and simplifying yields 

\begin{equation}
\begin{split}
&0=\frac{a_0^2T^8}{2}+a_0v_0T^7+(1+a_0(p_{end}-p_0))T^6+(20a_0^2v_0)T^5\\
&+(40a_0v_0+10a_0^2-717a_0^2/4)T^4+40a_0(p_{end}-p_0)T^3\\
&-697a_0v_0T^3+(20a_0(p_{end}-p_0)+20v_0+10a_0v_0)T^2\\
&-717(v_0+p_{end}a_0-a_0p_0)T^2-(1434v_0(p_{end}-p_0)T\\
&-717(p_0^2+p_{end}^2-2p_{end}p_0)
\end{split}
\label{closed_form_time}
\end{equation}

At this juncture, the other dimensions are lumped into the coefficients for finding the trajectory duration. The resulting eighth order polynomial is solved using the numerically stable Jenkins-Traub algorithm \cite{jenkins}\footnote{A C++ implementation of the algorithm is available at \href{https://github.com/ethz-asl/mav_trajectory_generation}{https://github.com/ethz-asl/mav\_trajectory\_generation}}.

The trajectory with the minimal average cost among all the real and positive solutions of Eq. \eqref{closed_form_time} is chosen after the coefficients are found.


The solution of Eq.\eqref{closed_form_time} may not result in real and positive roots. Therefore, an assumption of constant acceleration is used in to solve for a trajectory duration $\tilde{T}$. As the resulting trajectory does not account for the change in acceleration over the duration, a uniform distribution around the calculated trajectory duration $\tilde{T}$ to find the optimal trajectory. From this distribution, a set of trajectories are sampled. Out of these sampled trajectories, the trajectory that minimizes Eq.\eqref{equation} is used.

\begin{argmini}
{T}{\frac{1}{T}\int_{0}^{T}  \norm{u} ^2 +1 \hspace{1mm} dt}{}{}
\label{samplingobjective}
\end{argmini}


 Thus, minimum-time smooth trajectories are generated for every other robot(That is, for $M-1$) that transmits it's current state and desired state. Each individual trajectory is represented by $\textbf{x}_{pred}(t)$. However, this trajectory does not account for inter-robot interactions. Therefore, it is reshaped by a trajectory compensator.

\subsection{Trajectory Compensator}
The trajectory compensator presented in this work accounts for the spatial compensation over short term horizons by constructing an auxiallary polynomial similar in structure to Eq.\eqref{optimal_state}. These trajectories are represented by $\textbf{x}_{cmp}(t)$.

\begin{equation}
\begin{split}
p^*_{comp}(t)&= \gamma_1t^5+\gamma_2t^4+\gamma_3t^3+\gamma_4t^2+\gamma_5t+\gamma_6\\
v^*_{comp}(t)&=5\gamma_1t^4+4\gamma_2t^3+3\gamma_3t^2+\gamma_4t+2\gamma_5 \\
a^*_{comp}(t)&=20\gamma_1t^3+12\gamma_2t^2+6\gamma_3t+2\gamma_4
\end{split}
\label{compensator}
\end{equation}

To solve for the coefficients from Eq.\eqref{compensator}, a moving horizon with $K_{comp}$ samples of transmitted states $\Tilde{\textbf{x}}_{obs}$ and trajectories $\textbf{x}_{pred}$ from subsection \ref{jerk-smooth} is used.  Using this, a least squares problem that minimizes   
\begin{equation}
\Delta \textbf{x}_{obs}  = \textbf{x}_{cmp} - (\Tilde{\textbf{x}}_{obs} - \textbf{x}_{pred})    
\end{equation}
 over the horizon is used. Additionally constraints are introduced to ensure that compensated trajectory does not reposition the trajectory at the current state($K_{comp}^{\text{th}}$) and end position. This constraint is added by setting $\textbf{x}_{cmp}(0)=0$ and $\textbf{x}_{cmp}(T)=0$ . This results in:

\begin{argmini!}
  {\textbf{x}_{cmp}}{\sum_{i=0}^{K_{comp}-1} \Delta \textbf{x}_{obs}^T\Delta \textbf{x}_{obs}}{}{}
  \label{comp_objective}
  \addConstraint{\textbf{x}_{cmp}(0)}{=0}{}
  \label{comp_const}
  \addConstraint{\textbf{x}_{cmp}(T)}{=0}
  \label{comp_end_const}
\end{argmini!}

The equality constrained quadratic program is solved using its KKT matrix; which is a linear system. The linear system is solved using the sparse \textbf{LDL}$^\text{T}$ \cite{davis2005algorithm} implementation on Eigen \cite{eigenweb}. Based upon this, the predicted state trajectory of an individual other robot is represented by: 
\begin{equation}
    \textbf{x}_{obs}(t)= \textbf{x}_{pred}(t) + \textbf{x}_{cmp}(t)
\end{equation}   

The deviations between the predicted trajectory and the actual generated trajectory is shown in Fig \ref{visualise prediction}. 

\section{Trajectory optimization}
\label{traj optimization}
The trajectory optimization NLP is formulated as:
\begin{argmini!}
  {\textbf{\textit{x}}}{\int_{0}^{T} C_{dynm} +C_{lim} +\sum_{i=1}^{M-1} C_{collis}   \hspace{1mm} dt + K_t T^2  }{}{}
  \label{traj_objective}
  \addConstraint{\textbf{\textit{x}}(0)}{=\textbf{x}_0}{}
  \label{staring_const}
  \addConstraint{\textbf{\textit{x}}(T)}{=x_{end}}
  \label{end_const}
\end{argmini!}

Where $C_{dynam}$,$C_{collis}$ and $C_{lim}$ are costs for trajectory smoothness, collision with other robots and dynamic limits respectively. $K_t T^2$ is added to minimize the time taken along the trajectory. The trajectories are represented as time parameterized polynomials of order five to assuage representational mismatch with the predicted trajectory. 
Therefore, for each dimension, the trajectory is represented by:
\begin{equation}
p(t)= \sum_{j=0}^{5}\alpha_j t^j
\label{decision with degree}
\end{equation}
 This results in the decision variables of the optimization problem being:
 \begin{equation}
\mathcal{D} = [\alpha_0 \hspace{2mm} \alpha_1 \hspace{2mm} \alpha_2 \hspace{2mm} \cdots \alpha_{6N-1} \hspace{1mm} T]^T
\label{variables}
\end{equation}

Where the first six variables represent the polynomial coefficients of the first dimension, the second six the second dimension and so on. and $T$ represents the trajectory duration.




\subsection{Trajectory Smoothness}
To ensure that the generated trajectory is smooth, a smoothness objective is added. This objective is represented as:
\begin{equation}
C_{dynam}=Q_{dynm}\norm{\frac{d^{3}p(t)}{dt^{3}}}^2 
\label{derivative}
\end{equation}
Where,$Q_{dynm}$ is the weight for the smoothness. The smoothness cost (Eq.\eqref{derivative}) has a closed-form solution in terms of the optimization variables from Eq.\eqref{variables}.

\subsection{Collision Avoidance}
The exponential barrier function for avoiding collisions with other robots in the environment is:


\begin{equation}
C_{collis}=Q_{obs}c(p(t),p_{obs}(t))
\end{equation}
Where, $Q_{obs}$ is collision avoidance weighing parameter, 
\begin{equation}
c(p(t),p_{obs}(t)) = 2\frac{\sum_i^N p_i(t)-p^i_{obs}(t)(v_i(t)-v^i_{obs}(t))}{\exp^{K_p(d(p(t),p_{obs}(t)))}} 
\label{exponential collision barrier}
\end{equation}
$c(x(t))$is the cost of collision avoidance  and $d(p(t),p_{obs}(t))$ is the collision avoidance constraint \eqref{collisionconstraint}. 
The cost in \eqref{exponential collision barrier} can be analytically integrated. The cost i is one of the two-fold reasons for the addition of the relative velocity term $v_i(t)-v^i_{obs}(t)$ in the collision cost, the other being the increased weightage when robots move in opposite directions axially.




\subsection{Dynamic limits}
 Inspired by unconstrained optimization methods for multirotor collision avoidance \cite{burri2015real,2017real,gao2017gradient}, the limits are accounted for  using a different exponential barrier function based on soft constraints; thereby allowing for continuous-time limit verification while also not adding constraints. The barrier function is:
\begin{equation}
C_{lim}=\sum_{i=1}^3 Q_{lim}D(\frac{d^{i}p(t)}{dt^{i}})
\end{equation}
Where $Q_{lim}$ is the tuning weight for the dynamic limits,
\begin{equation}
D(x(t))=2x(t)\frac{dx(t)}{dt}\exp^{K_p(\norm{x(t)}^2-\tau_i^2)}
\end{equation}
Here $\tau_i$ is the maximum allowed magnitude of the $i^{\text{th}}$ derivative of the robot position and $\norm{.}$ is the Euclidean norm. Euclidean norm is used to account for coupled dynamic limits rather than each dimension individually. This is done as dynamic limits of robots are often coupled. The decoupling of the dynamic limits results in a conservative approximation. Limits on the robot's velocity, acceleration and jerk are applied. The cost is analytically integrated.

The NLP formed by Eq.\eqref{traj_objective},\eqref{staring_const},\eqref{end_const} cannot be proven to be convex due to the equality constraints in  Eq. \eqref{end_const}. Owing to the usage of soft constraints for collision avoidance (Eq.\eqref{exponential collision barrier}), collision-avoidance and dynamic limits' adherence cannot be guaranteed.
\subsection{NLP Solver}
\label{unconstrained solver}
To solve the NLP, the boundary value problem constructed by Eq.\eqref{staring_const},\eqref{end_const} is utilized to condense the optimization variables and convert it into an unconstrained NLP. Assuming, the Eq.\eqref{end_const} is fully defined, by substituting Eq. \eqref{staring_const} the constraint is represented(per dimension) as:

\begin{equation}
\begin{split}
p_{end}&= \alpha_1T^5+\alpha_2T^4+\alpha_3T^3+\frac{a(0)T^2}{2}+v(0)T+p(0)\\
v_{end}&=5\alpha_1T^4+4\alpha_2T^3+3\alpha_3T^2+a(0)T+v(0) \\
a_{end}&=20\alpha_1T^3+12\alpha_2T^2+6\alpha_3T+a(0)
\end{split}
\label{BVP}
\end{equation}

Rearranging it:

\begin{equation}
    \begin{bmatrix}
    \alpha_1 \\
    \alpha_2 \\
    \alpha_3
    \end{bmatrix} = A_{trsfrm}^{-1}\begin{bmatrix}
    p_{end} - (0.5a(0)T^2+v(0)T+p(0)) \\
    v_{end} - (a(0)T+v(0)) \\
    a_{end} - a(0)
    \end{bmatrix}
    \label{transformedc}
\end{equation}

with \begin{equation*}
A_{trsfrm}=\begin{bmatrix}
T^5 & T^4 & T^3 \\
5T^4 & 4T^3 & 3T^2 \\
20T^3& 12T^2& 6T
    \end{bmatrix}
\end{equation*}

Substituting Eq.\eqref{transformedc} in Eq.\eqref{traj_objective} transforms the optimization variables from Eq.\eqref{variables} to only $T$ but parametrized by $x_0$ and $x_{end}$. The final optimization problem is:

\begin{argmini}
  {T}{\int_{0}^{T} C_{dynm} +C_{lim} +\sum_{i=1}^{M-1} C_{collis}   \hspace{1mm} dt + K_t T^2  }{}{}
  \label{unconstrained}
\end{argmini}




\section{Results}
\label{results}
The algorithm was implemented in C++ using Robot Operating System(ROS). The algorithm was tested on a workstation with Intel Xeon E5 1630v5  processor, 64GB of RAM and a Nvidia Quadro M4000 GPU. The algorithm was verified with different flying robots of different sizes(Asctec Hummingbird\footnote{\href{http://www.asctec.de/uav-uas-drohnen-flugsysteme/asctec-hummingbird/}{http://www.asctec.de/uav-uas-drohnen-flugsysteme/asctec-hummingbird/}},Firefly\footnote{\href{http://www.asctec.de/uav-uas-drohnen-flugsysteme/asctec-firefly/}{http://www.asctec.de/uav-uas-drohnen-flugsysteme/asctec-firefly/}},Neo\footnote{\href{http://www.asctec.de/uav-uas-drohnen-flugsysteme/asctec-neo/}{http://www.asctec.de/uav-uas-drohnen-flugsysteme/asctec-neo/}}) in Gazebo using RotorS \cite{rotors}, a high fidelity multirotor simulator. Additionally, $\eta=3$. The constraints for the robots are given in Table \ref{parameters}


\begin{table}
\caption{parameters of the robots}
\label{parameters}
\begin{center}
\begin{tabular}{c|c|c|c|c}
\hline
\textbf{Robot} & \thead{\textbf{Size} $r_1$ \\ (m)} & \thead{\textbf{Vel} \\ (m/s)} & \thead{\textbf{Accel} \\ (m/s$^2$) } & \thead{\textbf{Jerk} \\ (m/s$^3$)} \\
\hline
Hummingbird & 0.4 &2 & 4 &20 \\
\hline
Firefly & 0.5 & 2 & 8 & 20 \\
\hline
Neo &0.6 & 4 & 12 & 60  \\
\hline
\end{tabular}
\end{center}
\end{table}

Rotary aerial robots' selection was guided by their differential flatness, fourth order dynamics \cite{mellinger2011minimum} and their nature as a complex, dynamic and 3D motion capability. Additionally, the low inertia and capacity to produce substantial torques allow them to be treated as a third order system. Tracking the generated trajectories was tested using a geometric $ SO(3)$ controller\cite{taecontroller} and model predictive control \cite{kamel2017linear}. The yaw motion of all the robots is kept free (zero) as any translational motion is invariant to rotation about yaw \cite{mellinger2011minimum,efficient2015quad}. The trajectories are re-planned at a frequency of 10Hz. Additionally, $v_{end}$ and $a_{end}$ from Section \ref{unconstrained solver} were set to zero.

The optimization problem in Eq.\eqref{unconstrained} is solved using \cite{nlopt}. The algorithm was tested with BOBYQA\cite{powell2009bobyqa}, MMA\cite{svanberg2002class},L-BFGS \cite{liu1989limited} and SLSQP\cite{sqp}. BOBYQA was the fastest while MMA and BOBYQA yielded similar performance. L-BFGS and SLSQP were slower. The first trajectory planning was initialized from the solution detailed in Section \ref{jerk-smooth}. Subsequent replanning was initialized by relatively scaling down the previous trajectory by the replanning rate.  

\subsection{Simulations}
Simulations were done to test the scalabilty and performance in dense environments of the proposed algorithm. In all the subsequent simulations, the occupancy was fixed by calculating the volume of each robot's ellipsoid and summing it up. Moreover, in the homogenous experiments, the robot used is AscTec Firefly. The computation time per replanning iteration is shown in Table \ref{computation time}

\begin{figure}
    \centering
    \includegraphics[width=0.48\textwidth]{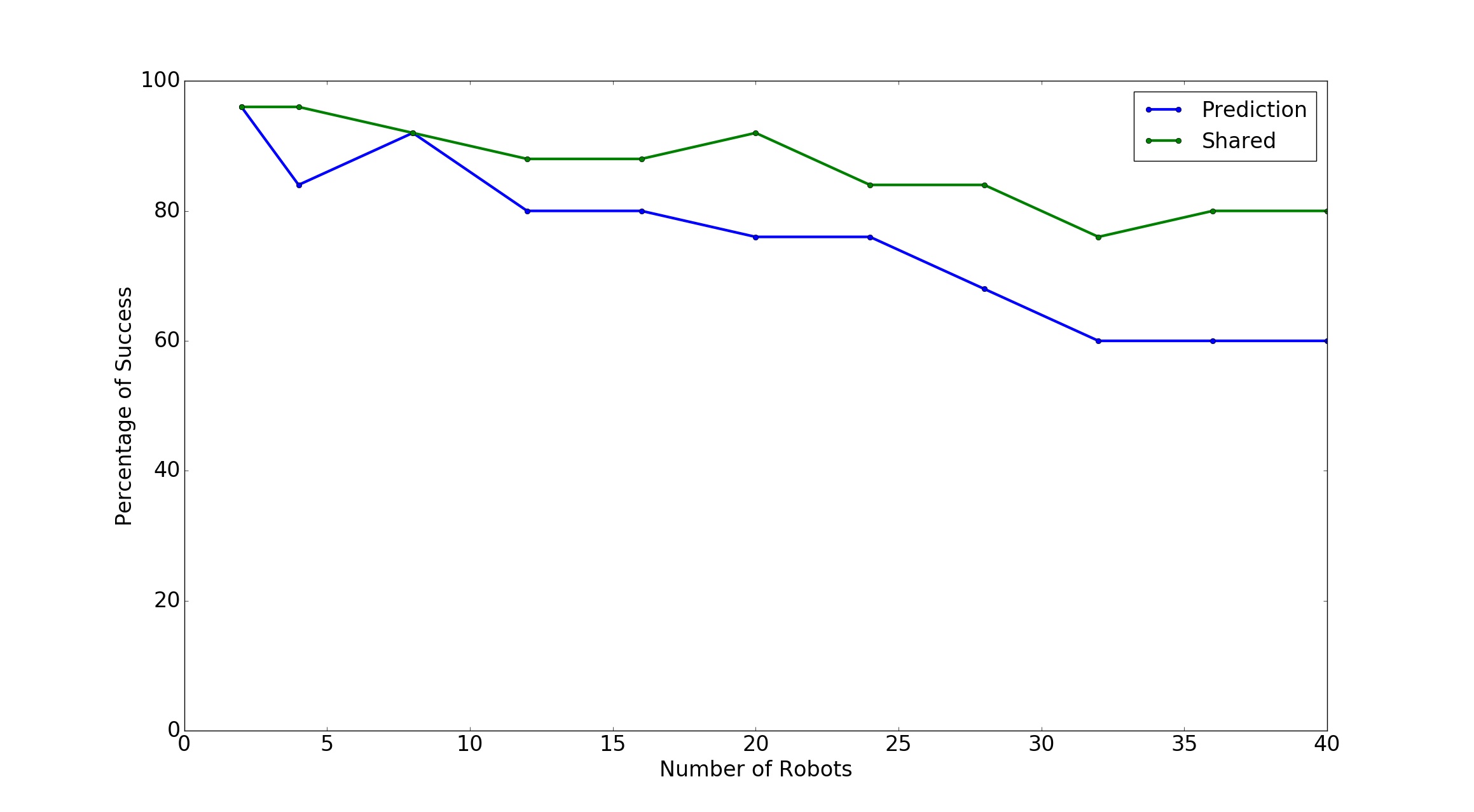}
    \caption{Performance of the proposed algorithm in simulation for increasing number of robots with a fixed occupancy of 20\%. For each quantity of robots, 25 random simulations were conducted}
    \label{Success homogenoeus}
\end{figure}

To test the scalability of the algorithm with homogeneous robots- keeping a fixed occupancy of 20\%(This translates to 5 agents/$m^3$ in the results from \cite{luis2019trajectory})- the number of robots were increased from 2 till 40. The scalability was tested is two ways: (i) The robots share their planned trajectories (ii) Trajectories are predicted. The ensuing performance is shown in Fig.\ref{Success homogenoeus}. The proposed trajectory optimization technique with robots sharing their trajectories shows a high success rate and accurate collision avoidance for upto forty robots in dense environments thereby showing the efficacy of the proposed condensation method and the objective functions. The algorithm with the robots predicting the trajectories shows a drop-off in performance as the number of robots scale up. This is an effect of the increasing inaccuracies in the trajectory prediction. The trajectories were tracked using linear model predictive control \cite{kamel2017linear}.

\begin{figure}
    \centering
    \includegraphics[width=0.48\textwidth]{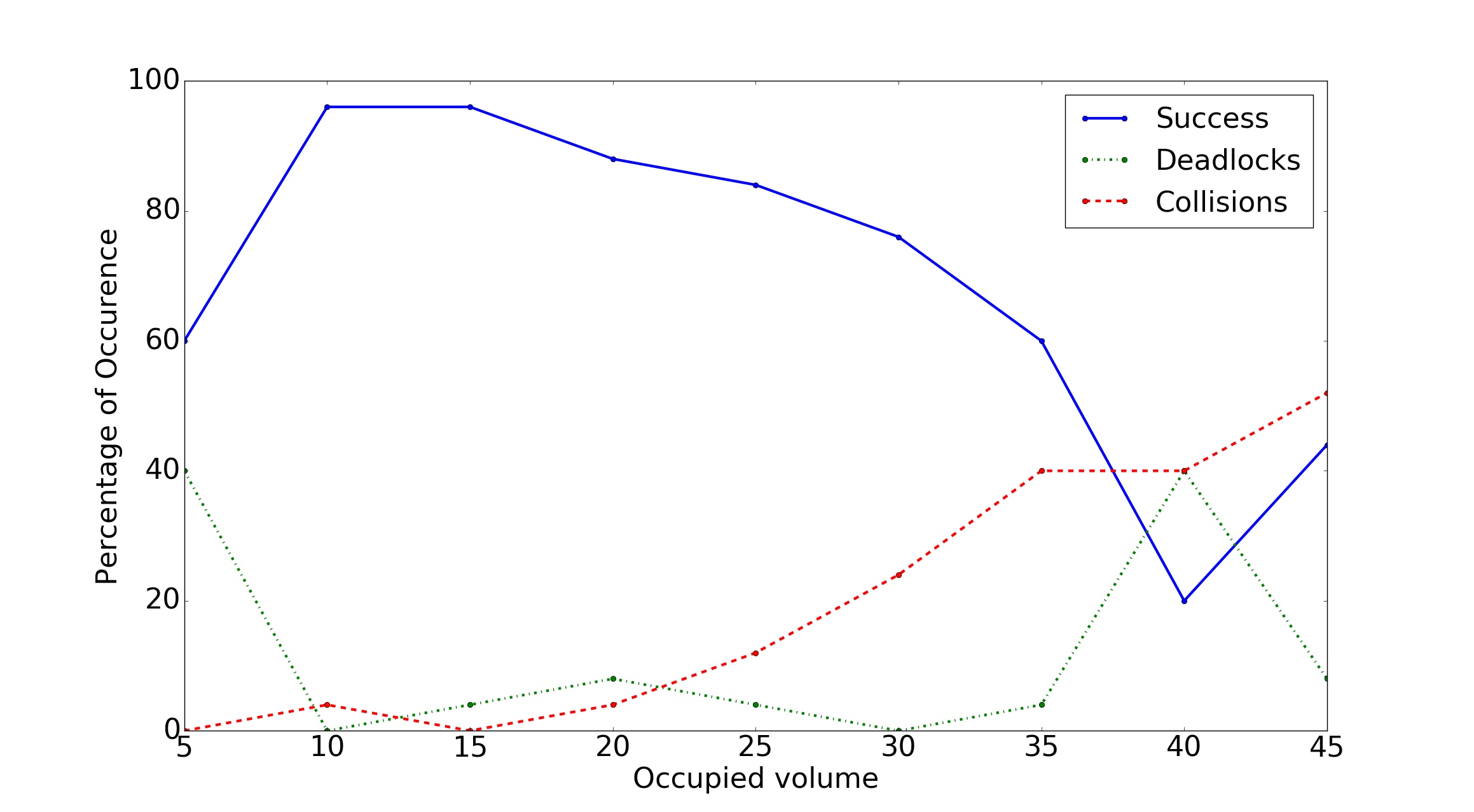}
    \caption{Performance of the proposed algorithm in simulation for increasing density of robots inside a fixed box of dimensions $4m$ x $4m$ x $2m$. For each quantity of robots, 25 random simulations were conducted}
    \label{Density}
\end{figure}

In a second set of simulations, whose results are shown in Fig.\ref{Density}, the performance with regards to increasing density of robots is tested. The robots are allocated end goal positions within a $4m$ x $4m$ x $2m$ box and tested with increasing number of robots. The number of deadlocks is high in sparse and small environments. This, we believe, is an artifact of the exponential barrier functions.

\begin{table}
\caption{Computation Time ($\mu$s)}
\label{computation time}
\begin{center}
\begin{tabular}{c|c|c|c|c}
\hline
 & \thead{\textbf{Mean}} & \thead{\textbf{Std}\\ \textbf{dev}} & \thead{\textbf{Min}} & \thead{\textbf{Max}} \\
\hline
Jerk-Smooth & 89.5 &12.91 & 75.24 &  126.32\\ 
\hline
Least-Squares & 3.9765 & 4.532 & 1.19 & 12.47 \\
\hline
Optimization & 3024.75 & 2407.93 & 998.06 & 24076.85 \\
\hline
\hline
\textbf{Total Time} & \textbf{3118.2265} & \textbf{2425.372}  \\
\end{tabular}
\vspace{1mm}
     
     {\raggedright \tiny{Please note that the Jerk-Smooth and Least-Squares computation time is for one single robot's prediction.  That is, if there are 8 other robots, then the computation time is going to be 716 for Jerk-Smooth and 31.812 for Least Squares  }\par}

\end{center}
\end{table}

Simulations have shown that the proposed algorithm is capable of generating collision-free trajectories in dense scenarios and is scalable to medium sized swarms of aerial robots. In contrast to other trajectory re-planning approaches, the proposed algorithm is computationally faster, continuous-time in nature and accounts for a higher order of dynamics. 

Another point to be noted is that owing to the indirect optimization of the polynomial coefficients, there were certain junctures when the planned trajectories dramatically veered. This effect was starkly visible especially when the nonlinear model predictive controller \cite{kamel2017linear} was used for tracking trajectories. 
 These dramatic changes were akin to an exaggerated version of random restarts used in local trajectory planning \cite{schulman2014motion} \cite{oleynikova2016continuous}. 

Paradoxically, despite the lack of boundaries for the robots' positions, the robots adhered to the box limits more consistently when the occupied volume was more dense. 
\section{Conclusion}
\label{conclusion}
A decentralized, local trajectory optimization algorithm for replanning continuous-time trajectories for multi-robot systems with third order dynamics was proposed. The algorithm predicts continuous time trajectories for other robots in the environment and utilizes those trajectories for planning collision-free trajectory for itself. Simulations were conducted with upto forty homogeneous robots at an occupancy of 20\%.  Building on this, two applications are envisioned for the proposed algorithm: (i) As part of trajectory generation pipeline to fix the optimal duration for the trajectory (ii) As part of a separate trajectory replanning pipeline.

In the future we would incorporate a better method to account for trajectory-tracking error. Additionally, we would look at integrating obstacles beyond ellipsoids and spheres. Improving upon the trajectory compensation by incorporating temporal compensation is another avenue for future work. Incorporating uncertainty into the prediction pipeline is something we would look to in the future.

\bibliographystyle{IEEEtran}
\bibliography{references,refer}

\addtolength{\textheight}{-16cm}

\end{document}